\documentclass[letterpaper]{article} % DO NOT CHANGE THIS
\usepackage{aaai20}  % DO NOT CHANGE THIS
\usepackage{times}  % DO NOT CHANGE THIS
\usepackage{helvet} % DO NOT CHANGE THIS
\usepackage{courier}  % DO NOT CHANGE THIS
\usepackage[hyphens]{url}  % DO NOT CHANGE THIS
\usepackage{graphicx} % DO NOT CHANGE THIS
\usepackage{graphicx}  % DO NOT CHANGE THIS
\newcommand{\citet}[1]{\citeauthor{#1}~\shortcite{#1}}
\newcommand{\citep}{\cite}

\usepackage{latexsym}
\usepackage{booktabs}
\usepackage{tabularx}
\usepackage{amsmath}
\usepackage{multirow}
\usepackage{csquotes}
\usepackage{enumitem}
\usepackage{url}

\usepackage{amssymb}% http://ctan.org/pkg/amssymb
\usepackage{pifont}% http://ctan.org/pkg/pifont

\title{Robust Named Entity Recognition with Truecasing Pretraining}

\author{Stephen Mayhew, Nitish Gupta, Dan Roth \\
  University of Pennsylvania \\
  Philadelphia, PA, 19104 \\
  {\tt \{mayhew, nitishg, danroth\}@seas.upenn.edu}}

\date{}

\begin{document}
\maketitle

\begin{abstract}

Although modern named entity recognition (NER) systems show impressive performance on standard datasets, they perform poorly when presented with noisy data. In particular, capitalization is a strong signal for entities in many languages, and even state of the art models overfit to this feature, with drastically lower performance on uncapitalized text. In this work, we address the problem of robustness of NER systems in data with noisy or uncertain casing, using a pretraining objective that predicts casing in text, or a \textit{truecaser}, leveraging unlabeled data. The pretrained truecaser is combined with a standard BiLSTM-CRF model for NER by appending output distributions to character embeddings. In experiments over several datasets of varying domain and casing quality, we show that our new model improves performance in uncased text, even adding value to uncased BERT embeddings.  Our method achieves a new state of the art on the WNUT17 shared task dataset.

\end{abstract}

\section{Introduction}

Modern named entity recognition (NER) models perform remarkably well on standard English datasets, with F1 scores over $90\%$ \cite{Lample2016NeuralAF}. But performance on these standard datasets drops by over 40 points F1 when casing information is missing, showing that these models rely strongly on the convention of marking proper nouns with capitals, rather than on contextual clues. Since text in the wild is not guaranteed to have conventional casing, it is important to build models that are robust to test data case.

\begin{figure}[t]
    \centering
    \includegraphics[scale=0.6]{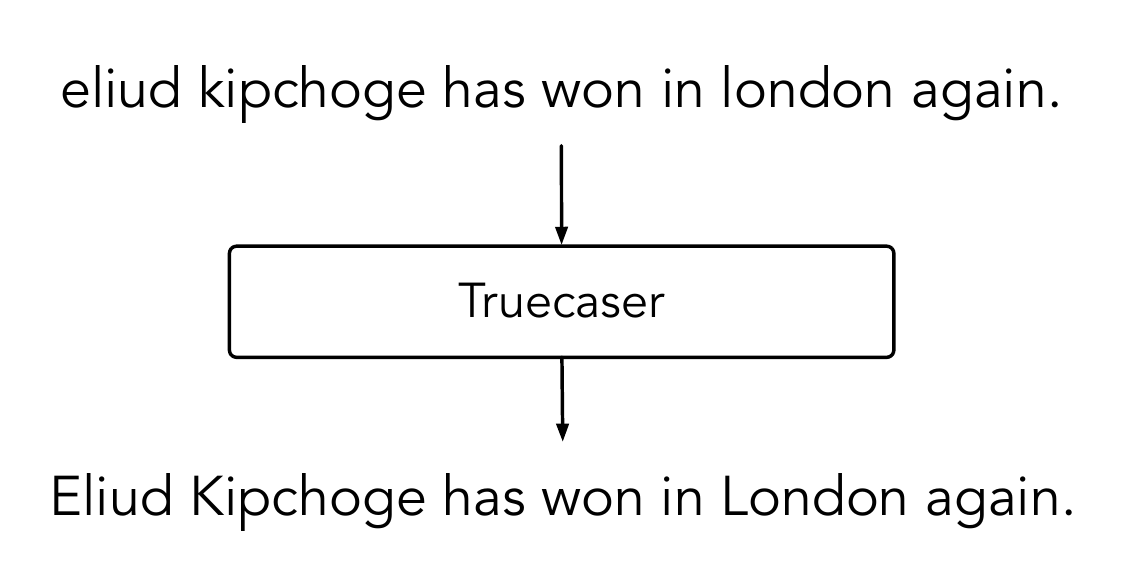}
    \caption{This figure shows the same sentence in two casing scenarios: on the top, uncased, on the bottom, truecased. Notice that the words marked as uppercase are also named entities. This paper explores how truecasing can be used in named entity recognition.}
    \label{fig:example}
\end{figure}

One possible solution to keep NER systems from overfitting to capitalization is to omit casing information. This would potentially lead to better performance on uncased text, but fails to take advantage of an important signal.

We build on prior work in truecasing \cite{susanto2016learning} by proposing to use a truecaser to predict missing case labels for words or characters (as shown in Figure \ref{fig:example}). Intuitively, since named entities are marked with capitals, a trained truecaser should improve an NER system. We design an architecture in which the truecaser output is fed into a standard BiLSTM-CRF model for NER. We experiment with pretraining the truecaser, and fine-tuning on the NER train set. When designing the truecaser, we suggest a preprocessing regimen that biases the data towards named entity mentions. 

In our experiments, we show that truecasing remains a difficult task, and further that although a perfect truecaser gives high NER scores, the quality of the truecaser does not necessarily correlate with NER performance. Even so, incorporating predicted casing information from the right truecaser improves performance in both cased and uncased data, even when used in conjunction with uncased BERT \cite{Devlin2018BERTPO}.  We evaluate on three NER datasets (CoNLL, Ontonotes, WNUT17) in both cased and uncased scenarios, and achieve a new state of the art on WNUT17.

\section{Truecaser Model}
\begin{figure}
    \centering
    \includegraphics[scale=0.5]{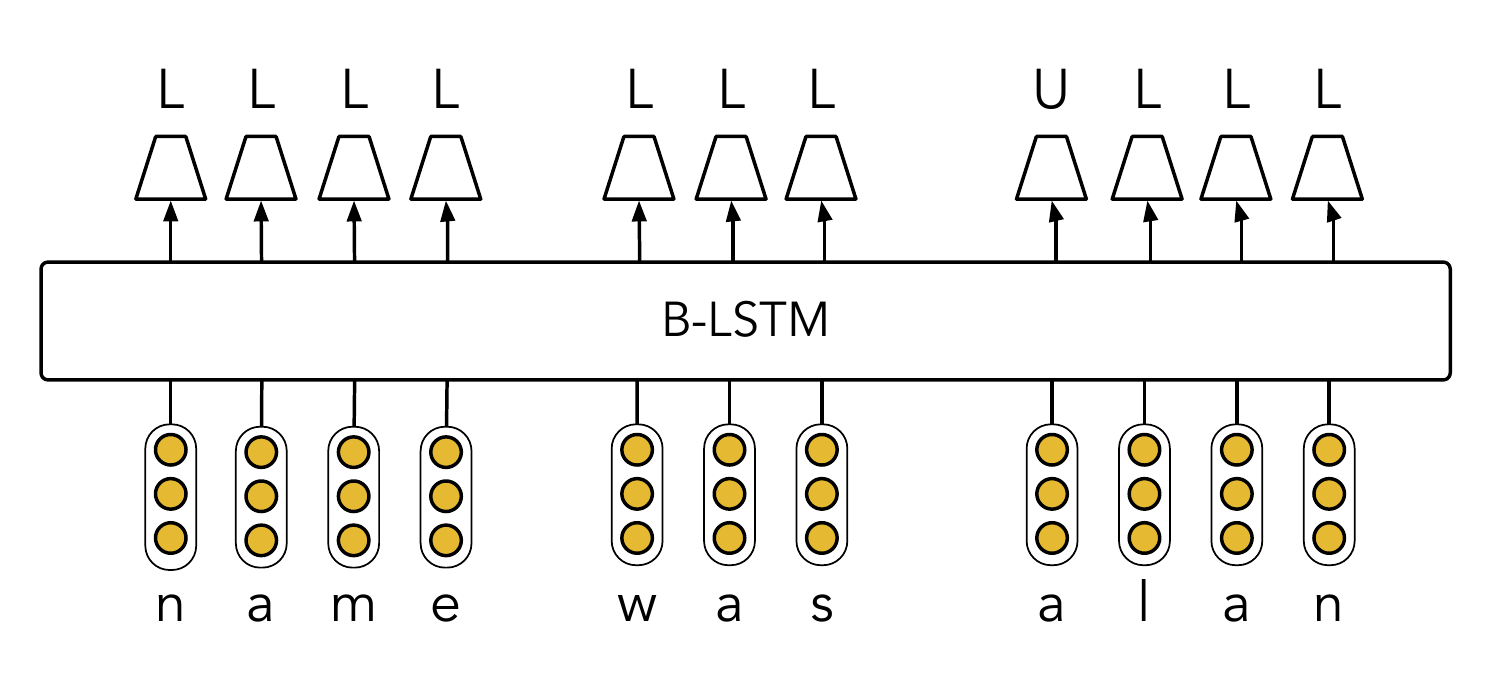}
    \caption{Diagram of our truecasing model. The input is characters, and the output hidden states from the BiLSTM are used to predict binary labels, `U' for upper case, and `L' for lower case. A sentence fragment is shown, but the B-LSTM takes entire sentences as input.}
    \label{fig:truecaser}
\end{figure}

A truecaser model takes a sentence as input and predicts case values (upper vs. lower) for all characters in the sentence.
We model truecasing as a character-level binary classification task, similar to \citet{susanto2016learning}. The labels, L for lower and U for upper, are predicted by a feedforward network using the hidden representations of a bidirectional LSTM (BiLSTM) that operates over the characters in a sentence. 
For example, in Figure~\ref{fig:truecaser}, the truecaser predicts ``name was Alan'' for the input ``name was alan''.
Formally, the output of this model is a categorical distribution $d_c$ over two values (true/false), for each character $c$,
\[ d_c = \text{softmax}(\mathbf{W}\mathbf{h}_c) \]

where $\mathbf{h}_c \in \mathbb{R}^H$ is the hidden representation for the $c$-th character from the BiLSTM, and $\mathbf{W} \in \mathbb{R}^{2\times H}$ represents a learnable feed-forward network. In addition to the parameters of the LSTM and the feedforward network, the model also learns character embeddings. 
Generating training data for this task is trivial and does not need any manual labeling; we lower-case text to generate input instances for the truecaser, and use the original case values as the gold labels to be predicted by the model. The model also uses whitespace characters as input for which the gold label is always lowercase (L) (not shown in Figure~\ref{fig:truecaser}).

\begin{figure}
    \centering
    \includegraphics[scale=0.5]{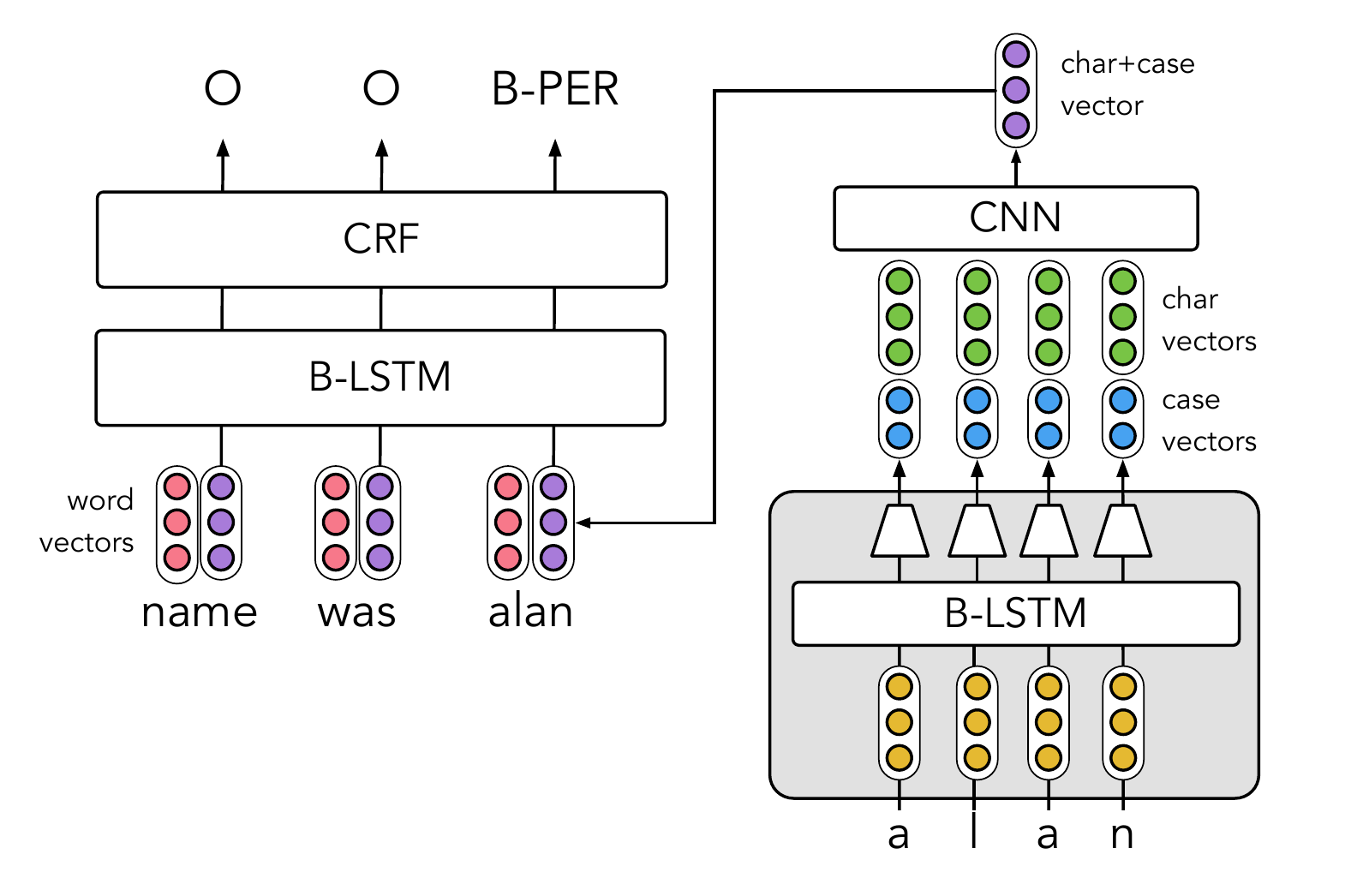}
    \caption{Diagram of our NER model. The left side shows a standard BiLSTM CRF model with word vectors (red) concatenated with character vectors (purple). The right side shows how the character vectors are created. The gray shaded area is our truecaser, with parameters frozen. This produces 2-dimensional case predictions (in blue) for each character. These are concatenated with learned character embeddings (in green) before going to a CNN for encoding.}
    \label{fig:ner}
\end{figure}

\section{Proposed NER Model}
In NER literature, the standard model is the BiLSTM-CRF. In this model, token embeddings are given as input to a BiLSTM, and the output hidden representations are in turn passed as features to a Conditional Random Field (CRF). The BiLSTM input embeddings are typically a concatenation of pretrained word embeddings (such as GloVe; ~\cite{Pennington2014GloveGV}) and encoded character embeddings. Given the popularity of this model, we omit a detailed description, and refer interested readers to prior work \cite{Chiu2016NamedER,Lample2016NeuralAF}.

Our proposed model augments the standard BiLSTM at the character level, appending truecaser predictions ($d_c$, from above) to each character embedding. Formally, let the input embedding for token $t$ be $x_t$.
\begin{align}
    x_t &= \begin{bmatrix}
          w_x \\
          f(c_1, c_2, ..., c_{|x_t|}) 
         \end{bmatrix}
  \end{align}

Here, word embeddings $w_x$ are concatenated with an embedding generated by $f(.)$, which is some encoder function over character embeddings $c_{1 : |x_t|}$ representing characters in $x_t$. This is usually either a BiLSTM, or, in our case, a convolutional neural network (CNN). Even if the word vectors are uncased, casing information may be encoded on the character level through this mechanism. 

For any $x_t$, our model extends this framework to include predictions from a truecaser for all characters, as a concatenation of character embedding $c_i$ with the corresponding truecaser prediction $d_i$.
\begin{align}
    v_i &= \begin{bmatrix}
          c_i \\
          d_i
         \end{bmatrix}
  \end{align}

Now, the input vector to the BiLSTM is:
\begin{align}
    x_t &= \begin{bmatrix}
          w_x \\
          f(v_1, v_2, ..., v_{|x_t|}) 
         \end{bmatrix}
  \end{align}

In this way, each character is now represented by a vector which encodes the value of the character (such as `a', or `p'), and also a distribution over its predicted casing. This separation of character and casing value has also been used in \citet{Chiu2016NamedER} and \citet{collobert2011natural}. The distinction of our model is that we explicitly use the predicted casing value (and not the true casing in the data). The reason for using predicted casing is to make the NER model robust to noisy casing during test time. Also, an NER system using gold-casing while training might overfit to quirks of the training data, whereas using predicted casing will help the model to learn from its own mistakes and to better predict the correct NER tag.

A diagram of our model is shown in Figure~\ref{fig:ner}. We refer to the two-dimensional addition to each character vector as ``case vectors." Notice that there are two separate character vectors present in the model: those that are learned by the truecaser and those that are learned in the NER model (char vectors). We chose to keep these separate because it's not clear that the truecaser character vectors encode what the NER model needs, and they need to stay fixed in order to work properly in the truecaser model. 

In our experiments, we found that it is best to detach the truecaser predictions from the overall computation graph. This means that the gradients propagated back from the NER tag loss do not flow into the truecaser. Instead, we allow parameters in the truecaser to be modified through two avenues: pretrained initialization, and truecaser fine-tuning. These avenues are orthogonal and can operate independently of each other or together. In one of our experiments, we try all combinations. 

\begin{table}[]
    \centering
    \begin{tabular}{llll}
    \toprule
        Dataset & Train & Dev & Test  \\
        \midrule
        Wikipedia & 2.9M & 294K & 32K \\
        Common Crawl & 24M & 247K & 494K \\  
        \bottomrule
    \end{tabular}
    \caption{Number of tokens in the train/dev/test splits for Wikipedia (Wiki) and Common Crawl (CC).}
    \label{tab:truecaser_data_statistics}
\end{table}

\section{Experimental Setup}
Having described our models, now we move to the experimental setup. 

\subsection{Truecaser data}

Supervision for a truecaser comes from raw text, with the only requirement being that the use of capitals follow certain desirable standards. We train on two different datasets: the Wikipedia dataset (Wiki) introduced in \citet{coster2011learning} and used in \citet{susanto2016learning}, and a specially preprocessed large dataset from English Common Crawl (CC).\footnote{\url{commoncrawl.org}} Statistics for each dataset are found in Table \ref{tab:truecaser_data_statistics}.

\begin{table}[]
\small
    \centering
    \begin{tabular}{p{0.45\textwidth}}
        \toprule
        Example sentences  \\
        \midrule
        `` the investigation is still ongoing , " McMahon said . \\
        he 's a Texas genius . \\
        Rothbart visits at 6 p.m. next thursday for a talk . \\
        has Smith made a decision ? \\
        in the Seventies , you would have written Britain off . \\
         \bottomrule
    \end{tabular}
    \caption{Example sentences from the Common Crawl truecasing training data. Notice that the first word is capitalized only for names, ``thursday'' is lowercased, and ``Seventies" is a non-named entity which was not caught by the lower-casing rules.}
    \label{tab:tc_example_sents}
\end{table}

We used the Wiki dataset as is, but modified the CC dataset in a few important ways. Since our ultimate goal is to train NER models, we applied rules with the aim of leaving only named entities capitalized. We did this in two ways: 1) convert the first word of each sentence to its most common form, and 2) lowercase certain words which are commonly capitalized, but rarely named entities. 

To accomplish the first, we used scripts from the moses package \cite{koehn2007moses} to collect casing statistics for each word in a large corpus, then replace the first word of each sentence with the most common form.\footnote{In a naming clash, the moses script is called a `truecaser', even though it only touches sentence-initial words.} For example, this sentence begins with `For', which is more likely to take the form `for' in a large corpus.

To accomplish the second preprocessing step, we briefly examined the training data, and applied rules that lower-cased a small number of words. For example, this list includes titles (Mr., Senator, Archbishop), month/day names (January, Thursday, although not April, or May, as these may well be person names), and other conventionally capitalized words which are not entities (GMT, PM, AM). These rules covered many commonly capitalized non-entities, but are not comprehensive.

Further, in order to avoid keeping titles or other upper cased sentences, we removed all sentences in which the ratio of capitalized words to total words exceeded 20\%. This removed such misleading sentences as ``Man Bites Dog In Pajamas." Table \ref{tab:tc_example_sents} shows some example sentences from the CC training data. 

\begin{table}[]
    \centering
    \begin{tabular}{lrrr}
    \toprule
    Dataset & Train & Dev & Test \\
    \midrule
    \multirow{2}{*}{CoNLL2003} & 203,621 & 51,362 & 46,435 \\
    & 23,499 & 5,942 & 5,648 \\
    \cmidrule{2-4}
    \multirow{2}{*}{Ontonotes} & 1,088,503 & 147,724 & 152,728 \\
    & 81,829 & 11,066 & 11,257 \\
    \cmidrule{2-4}
    \multirow{2}{*}{WNUT17} & 62,730 & 15,733 & 23,394 \\
    & 1,975 & 836 & 1,079 \\
    \bottomrule
    \end{tabular}
    \caption{Data sizes of the NER datasets. In each group, the top row shows number of tokens, the bottom row shows number of entity phrases.}
    \label{tab:ner_data_statistics}
\end{table}

\subsection{NER Data}
We experiment on 3 English datasets: CoNLL 2003 \cite{Sang2003IntroductionTT}, Ontonotes v5 \cite{HMPRW06}, WNUT17 \cite{derczynski2017results}. Data statistics are shown in Table \ref{tab:ner_data_statistics}.

CoNLL 2003 English, a widely used dataset constructed from newswire, uses 4 tags: Person, Organization, Location, and Miscellaneous. 

Ontonotes is the largest of the three datasets, and is composed of 6 diverse sub-genres, with a named entity annotation layer with 17 tags. We use the v5 split from \citet{pradhan2012conll}.

WNUT17 is the dataset from the shared task at the Workshop for Noisy User-generated Text 2017 (WNUT17), with 6 named entity tags. Since the focus of the shared task was on \textit{emerging entities}, the dev and test sets are considerably different from the train set, and have a distinctly low entity overlap with train.

\subsection{Training Details}
When setting up experiments, we always assume that we have access to cased training data, and the uncertainty arises in the casing of the test data. We suggest that there are only two casing scenarios for test data: 1) test data is cased correctly with high probability, (this may be the situation if the text comes from a reputable news source, for example) and 2) test data in which there is some doubt about the quality of the casing. In such a case, we argue that the text should be all lowercased as a preprocessing step. 

These two scenarios will guide our experiments. In the first scenario, high-likelihood cased test data, we will train models on cased text, and evaluate on cased text, as well as uncased text. This simulates the situation in which lower-cased text is given to the model. Since there is a casing mismatch between train and test, we would expect these numbers to be low. In the second scenario, we target known lowercased text and all training data is lowercased. 

In all experiments, we used a truecaser trained on the Common Crawl data, with character embedding of size 50, hidden dimension size 100, and dropout of 0.25. One trick during training was to leave 20\% of the sentences in their original case, effectively teaching the truecaser to retain casing if it exists, but add casing if it doesn't. We refer to this as \textit{pass-through truecasing}. 

For the NER model, we used character embeddings of size 16, hidden dimension size 256, and GloVe 100-dimensional uncased embeddings \cite{Pennington2014GloveGV}.\footnote{\url{nlp.stanford.edu/projects/glove/}} Even though the word embeddings are uncased, the character embeddings retain case.

For the WNUT17 experiments, we used 100-dimensional GloVe twitter embeddings, and 300-dimensional embeddings from FastText,\footnote{\url{fasttext.cc}} trained on English Common Crawl. 

For BERT, we used the model called {\verb bert-base-uncased } as provided by HuggingFace.\footnote{\url{github.com/huggingface/pytorch-pretrained-BERT/}} All experiments used AllenNLP \cite{Gardner2017AllenNLP}.

\section{Experiments and Results}

This section describes our experiments and results, first for the truecaser, then for the NER model.

\subsection{Truecaser}

We train our truecaser on each of the two datasets (Wiki and CC), and evaluate on the corresponding test sets, as well as on the test sets of NER datasets. Further, we train the truecaser on NER training sets (without using the NER labels), and evaluate respectively. Performance is shown in Table \ref{tab:truecaser_results}, using character-level F1, with U as the positive label. 

The highest scores for each test set are obtained with models trained on the corresponding training set. Given the difference in training data preprocessing, the performance of Wiki and CC are not comparable. All the same, a pattern emerges between the two truecasers: the highest performance is on Ontonotes, then CoNLL, then WNUT17. In fact, this pattern holds even in the bottom section, where the NER training set was used for training.

\begin{table}
    \centering
    \begin{tabular}{llcc}
    \toprule
        Train & Test &~~~& F1  \\
        \midrule
        Wiki & Wiki && 91.8 \\
        CC & CC && 81.9 \\
        \midrule
        Wiki & Ontonotes &&  79.9 \\
        Wiki & CoNLL && 64.3  \\
        Wiki & WNUT17 &&  47.0 \\
        \midrule
        CC & Ontonotes &&  61.7 \\
        CC & CoNLL && 57.3 \\
        CC & WNUT17 && 30.6 \\
        \midrule
        Ontonotes & Ontonotes && 86.3 \\
        CoNLL & CoNLL && 81.9 \\
        WNUT17 & WNUT17 && 52.3 \\
         \bottomrule
    \end{tabular}
    \caption{Truecasing performance on the character level (precision, recall, F-measure). In each of the three bottom sections, there is a descending pattern of F1 scores from Ontonotes, CoNLL, WNUT17. This suggests decreasing observance of capitalization conventions.}
\label{tab:truecaser_results}
\end{table}

Given that the pattern holds not only across truecasers from separate data, but also when training on in-domain training sets, this suggests that the explanation lies not in the power of the truecaser used to predict, but in the consistency (or inconsistency) of the data. 

Since capitalization is just a convention, truecasing is not well-defined. For example, if in this sentence, THESE WORDS were capitalized, it would not be so much ``wrong" as unconventional. As such, the rankings of scores should be understood as a measure of how much a particular dataset follows convention. These observations ring true with inspection of the data. Ontonotes, with some exceptions, follows relatively standard capitalization conventions. CoNLL is slightly less standard, with a combination of headlines and bylines, such as ``RUGBY UNION - CUTTITTA BACK FOR ITALY", and tables summarizing sporting events. In a few memorable cases, capitalization is repurposed, as in: 

\begin{displayquote}
League games on Thursday ( home team in CAPS ) : Hartford 4 BOSTON 2
\end{displayquote}

No truecaser could be reasonably expected to correctly mark capitals here.

Finally, WNUT17, the twitter dataset, has the lowest scores, suggesting the lowest attention to convention. Intuitively, this makes sense, and examples from the training set confirm it: 
\begin{itemize}[noitemsep]
    \item HAPPY B . DAY TORA *OOOOOOOOO* ~
    \item ThAnK gOd ItS fRiDaY !!
    \item im thinking jalepeno poppers tonight :] ]
    \item Don Mattingly will replace Joe Torre as LA Dodgers manager after this season\
\end{itemize}

Ultimately, it's important to understand that truecasers are far from perfect. It may seem like a simple task, but the many edge cases and inconsistencies can lead to poor performance.

\begin{table}[]
    \centering
    \begin{tabular}{llrr}
        \toprule
        Model & Init & NER F1 & Char F1  \\
        \midrule
        \multirow{5}{*}{\shortstack[l]{BiLSTM-CRF\\~~+GloVe}} & None & 87.4 & 0.0 \\
        &Gold & 90.4 & 100.0 \\
        \cmidrule{2-4}
        &Wiki &  87.4 & 63.7 \\
        &CC & \textbf{88.3} & 58.6 \\
        &CoNLL & 85.3 & 81.9 \\
        \bottomrule
    \end{tabular}
    \caption{CoNLL2003 testb performance of BiLSTM-CRF+GloVe uncased with different pretraining initializations, no auxiliary loss. NER F1 does not correlate with truecaser character-level F1.}
    \label{tab:ner_conll_diff_init}
\end{table}

\begin{table*}[]
    \centering
    \begin{tabular}{p{2.5cm}lccccccc}
      && \multicolumn{2}{c}{CoNLL} & \multicolumn{2}{c}{Ontonotes} & \multicolumn{2}{c}{WNUT17} & \\
    \toprule
        Train Case & Method & C & U & C & U & C & U & $\delta$ \\
        \midrule
        \multirow{7}{*}{Cased} & \cite{Lample2016NeuralAF} & 90.9  & -- & -- & -- & -- & -- \\
        &\cite{Chiu2016NamedER} & 91.6$^\dagger$  & -- & 86.3$^\dagger$ & -- & -- & -- \\
        &\cite{clark2018semi} & 92.6 & -- & 88.8 & -- & -- & -- \\
        &\cite{Devlin2018BERTPO} & 92.8  & -- & -- & --& -- & -- \\
        &\cite{akbik-etal-2018-contextual} & -- & -- & -- & -- & 46.0$^*$ & -- \\ 

        & BiLSTM-CRF+GloVe uncased & 90.2 & 24.3 & 87.3 & 83.5 & 41.0 & 15.7& \\
        \addlinespace[0.5mm]
        & ~~+truecaser & \textbf{90.3} & \textbf{84.5} & \textbf{87.0} & \textbf{81.1} & \textbf{43.7} & \textbf{30.2} & +12.5 \\
        \midrule
        \multirow{1}{*}{Cased+Uncased} & Data augmentation (Mayhew et al. 2019)& 90.4 & 87.9 & 87.2 & 84.3 & 38.6 & 34.8   \\
        \midrule
        \multirow{6}{*}{Uncased} &\cite{aguilar2018modeling} & -- & --  & -- & -- & -- & 45.6 \\
         & GloVe + Gold case vectors & --& 90.4 & --&86.9 & --&42.8 \\
        \addlinespace[0.5mm]
        & BiLSTM-CRF+GloVe uncased & --& 87.3 & --&83.4 & --&41.1 \\
        &~~+truecaser & --& 88.3 & --&84.4 & --&43.2 & +1.4 \\
        
        &BiLSTM-CRF+BERT uncased & --&91.0 & --& \textbf{88.1} & --&46.1 \\
        &~~+truecaser & --& \textbf{91.2} &--& \textbf{88.1} &--& \textbf{46.9} & +0.3 \\
        \bottomrule
    \end{tabular}
    \caption{F1 scores on cased (C) and uncased (U) test data. Models are trained on cased text in the upper section, and uncased text in the lower section. +truecaser means we include our pretrained two-dimensional character embeddings without fine-tuning. $^\dagger$uses additional gazetteer features. $^*$our run using their code, training only on train data. The $\delta$ columns shows average performance improvements of adding the truecaser.}
    \label{tab:results}
\end{table*}

\subsection{Truecaser Initialization for NER} 
We would like to know which truecaser training data performs best for NER. We would also like to understand if there is a correlation between truecaser and NER performance.

To measure this, we train a standard BiLSTM-CRF with GloVe embeddings on uncased text, using several different truecaser pretraining initializations, shown in Table~\ref{tab:ner_conll_diff_init}. As baseline and ceiling respectively, we use No initialization (None) and perfect predictions (Gold). Further, we compare using a truecaser trained on Wiki, trained on CC, and trained on CoNLL training data directly.

With \textbf{no initialization}, the model is poor, and with \textbf{perfect initialization} (gold predictions), the model has performance comparable to models trained on cased text. However, the story with learned initializations shows that higher Char F1 does not lead to higher NER F1.

This may seem counter-intuitive, but consider the following. There exists a truecaser that capitalizes the first word of each sentence, and every occurrence of `Mrs.' and `I'. Such a truecaser would have reasonably high Char F1, but poor NER performance. Conversely, there also exists a truecaser that capitalizes every named entity and nothing else, it would clearly be useful for NER, but would perform poorly on truecaser metrics.

In this light, if you can't build a perfect truecaser, then targeting entities is the next best option. This validates our decision to preprocess the CC truecaser training data to more closely track with named entities. With this intuition, and the results from this experiment, we chose to use CC in all further experiments.

The \textbf{CoNLL-trained truecaser} achieves high Char F1, but hurts the NER performance. We suspect that the truecaser learns to capitalize certain easy words (first word of sentence, day names) and to memorize names in the training data, and the model learns to rely on this signal, as in regular cased data. But at test time, the remaining 19\% F1 consists of names that the truecaser fails to correctly capitalize, thus misleading the NER model.

\subsection{Main Results} 
Using the CC truecaser, we experiment in several settings on the three NER datasets. All results are shown in Table \ref{tab:results}.

\paragraph{Cased Training Data} In the first setting (top section of the table), we train on cased NER data, resulting in standard performance on the cased data, but a severe drop in performance on the uncased data (from $90.2$ F1 to $24.3$ F1). This is the performance one might see when using a typical NER model in the wild. Adding our truecaser representations, we maintain scores on cased test data, and significantly improve performance on uncased data; by an average of $12.5$ F1. 

The middle section of Table \ref{tab:results} shows results from a data augmentation approach proposed in \citet{mayhew-etal-2019-ner}. In this approach, we simply concatenate cased training data with a lowercased copy of itself. This leads to strong performance on the average of cased and uncased outputs, although the outcome in the WNUT17 is somewhat degraded. 

\paragraph{Uncased Training Data} In the second experiment (lower half of Table~\ref{tab:results}), we address the scenario in which the casing of the test data cannot be trusted, leading us to lower-case all test data as a pre-processing step.

In theory, appending gold case labels to the character vectors (as in \citet{Chiu2016NamedER}, for example) should result in  performance comparable to training on cased data. This is the case for CoNLL and Ontonotes, as shown in the \textit{GloVe + Gold case vectors} row.

We experiment with two different embeddings: using GloVe uncased embeddings as before, or using BERT uncased embeddings. In each setting, we train models with and without the truecaser representations.

The first experiment with GloVe is equivalent to training on lower-cased text only, which is about 3 points lower than the cased version. This drop is much lower than the performance gap seen when training on cased data, but tested on uncased data. When we use our truecaser representations in this model, we see a consistent improvement across datasets, with an average of 1.4 points F1. This goes to show that even when training on lower-cased data, using the predicted truecasing labels helps the model perform better on uncased text.

Although the data augmentation approach is effective for average performance on cased and uncased data, if the target data is known to be lowercased (or if one decides to lowercase it because of uncertain casing), then the approach in this section has the best performance. 

We gathered scores for each sub-genre in Ontonotes, displayed in Table~\ref{tab:ontonotes}. Performance increases in every subsection, with the greatest gains in the broadcast conversation (bc) subsection (1.9) and the smallest in the telephone conversation (tc) subsection (0.3).

When using BERT uncased embeddings, even without the truecaser, the performance is better than a cased model without BERT, which shows that BERT is a strong contextual token embedder as compared to a BiLSTM with GloVe embeddings. Using the truecaser along with BERT, the model performance still improves (average of 0.3 points F1), showing that the truecaser is able to provide the model with complementary information that BERT does not capture.

\paragraph{Fine-Tuning and Pretraining} So far, we have used a pretrained truecaser with no fine-tuning. However, we can fine-tune the truecaser on training data with case labels, when available. 

 Table~\ref{tab:wnut_experiments} on WNUT17 shows that fine-tuning a pretrained truecaser substantially increases truecasing performance from $30.6$ F1 to $52.3$ F1. Though, this increase does not translate to an increase in NER performance, perhaps because of the domain mismatch in the pretrained truecaser's training data and WNUT17.
 
When we do not initialize the truecaser with pretrained parameters, but train the truecaser from scratch in combination with the NER objective, although the truecasing performance is not the highest, the NER performance improves greatly, achieving state of the art performance on WNUT17. In this case, during the initial iterations, the NER model receives random truecaser predictions, encouraging it to discriminate using the context. As the truecaser improves, the NER model now receives casing predictions helpful for learning, leading to improved performance.

\section{Related Work}
Our work uses truecasing as a pretraining objective, with the specific goal of improving NER. In this section, we discuss prior work in each area. 

\paragraph{Truecasing}  Early work in truecasing \cite{brown2001capitalization} was motivated by the prevalence of tasks that produced ``case deficient'' outputs, such as closed-caption TV, and automatic speech recognition (ASR) \cite{kubala1998named}. The proposed solution was a mix of heuristics (capitalizing single letters followed by a period), learned rules (frequency tables of commonly capitalized words), and contextual clues learned from running an NER system over well-cased data. 

\begin{table}[]
    \centering
    \begin{tabular}{lrrr}
        \toprule
        Test Set & F1 & F1 +truecaser & $\delta$ \\
        \midrule
        All (micro) & 83.4 & \textbf{84.4} & 1.0 \\
        \midrule
        bc & 81.8 & \textbf{83.7} & 1.9 \\
        bn & 87.5 & \textbf{88.6} & 1.1 \\
        mz & 79.5 & \textbf{81.0} & 1.5 \\
        nw & 85.8 & \textbf{86.6} & 0.8 \\
        tc & 69.6 & \textbf{69.9} & 0.3 \\
        wb & 76.1 & \textbf{76.6} & 0.5 \\
         \bottomrule
    \end{tabular}
    \caption{Scores on uncased Ontonotes test set from an uncased model, broken down by sub-genre. The leftmost column is a standard BiLSTM-CRF, the middle column is our proposed approach, and the rightmost column is the difference. In all cases, the truecaser outperforms the original model, with the greatest improvements in bc (broadcast news) and mz (magazine).}
    \label{tab:ontonotes}
\end{table}

\begin{table}[]
    \centering
    \begin{tabular}{lrr}
    \toprule
        TC Train & NER F1 & Char F1  \\
    \midrule
        Fixed pretrained &  46.9 & 30.6 \\
        Fine-tuned pretrained & 46.3 & \textbf{52.3} \\
        \addlinespace[0.5mm]
        From scratch & \textbf{47.7} & 36.2 \\
    \bottomrule
    \end{tabular}
    \caption{Results on WNUT17 with BERT uncased, trained and tested on uncased (U) data, varying the Truecaser (TC) training paradigm.}
    \label{tab:wnut_experiments}
\end{table}

Further innovations include lattice-based decoding with language models \cite{lita2003truecasing}, and truecasing as sequence tagging \cite{chelba2006adaptation}, with applications to machine translation \cite{wang2006capitalizing} and social media \cite{nebhi2015restoring}. Later, \citet{susanto2016learning} study the task with character-based representations and recurrent neural networks. 

\paragraph{Pretraining Objectives} In recent years, several works have shown how models trained over large amounts of raw text can produce powerful contextual representations \cite{Devlin2018BERTPO,Peters2018DeepCW}. In most works, the training objective is language modeling, or predicting a masked word given context. Our work can be seen as pretraining with a truecasing objective.

\paragraph{Named Entity Recognition} Named Entity Recognition (NER) is the task of identifying and classifying named entity mentions, such as persons, organizations, and locations. NER is typically modeled as a sequence tagging problem, in which each word is given a named entity tag, often using BIO encoding \cite{RamshawMa96} to mark phrases.

Early models of NER used linear methods, including classifiers with Conditional Random Fields (CRF) \cite{FinkelMa09}, and weighted averaged perceptron \cite{RatinovRo09}. Neural models based on BiLSTM-CRF have since shown strong performance, and have consequently become ubiquitous \cite{Chiu2016NamedER,Lample2016NeuralAF}.

When using the BiLSTM-CRF model, it is common to include character embeddings, either encoded with a convolutional neural network (CNN) \cite{Chiu2016NamedER}, or an LSTM \cite{Lample2016NeuralAF}. In some cases, notably \cite{Chiu2016NamedER} and \cite{collobert2011natural}, casing is included as an explicit feature in the character and word embeddings. Our work is similar to this, except we predict the casing.

In recent years, contextual representations such as ELMO \cite{Peters2018DeepCW} and BERT \cite{Devlin2018BERTPO} have proven to be remarkably successful for NER.

While there is little work that targets robustness of general NER models, there has been work on NER in noisy domains like twitter \cite{ritter2011named}, and several Workshops on Noisy User-generated Text (WNUT) \cite{strauss2016results,derczynski2017results}. In particular, \cite{aguilar2018modeling} target the WNUT17 task, achieving strong results using a phonetic representation to model their text (with no capitalization, incidentally), and including multitask objectives. Recent work has suggested data augmentation as a solution \cite{mayhew-etal-2019-ner,bodapati-etal-2019-robustness}.

\section{Conclusions}
We have shown how pretraining with a truecasing objective can improve the robustness of a named entity recognition system to in both cased and uncased test scenarios. Our experiments with varied types of truecasing training data give insights into best practices for preprocessing. Finally, we have demonstrated that BERT uncased representations are helpful for lowercased NER, but can also be extended with our techniques.

\section{Acknowledgments}
This work was supported by Contracts HR0011-15-C-0113 and HR0011-18-2-0052 with the US Defense Advanced Research Projects Agency (DARPA). Approved for Public Release, Distribution Unlimited. The views expressed are those of the authors and do not reflect the official policy or position of the Department of Defense or the U.S. Government.

\bibliography{ccg,cited,mybib}

\begin{thebibliography}{}

\bibitem[\protect\citeauthoryear{Aguilar \bgroup et al\mbox.\egroup
  }{2018}]{aguilar2018modeling}
Aguilar, G.; Monroy, A. P.~L.; Gonz{\'a}lez, F.; and Solorio, T.
\newblock 2018.
\newblock Modeling noisiness to recognize named entities using multitask neural
  networks on social media.
\newblock In {\em Proceedings of the 2018 Conference of the North American
  Chapter of the Association for Computational Linguistics: Human Language
  Technologies, Volume 1 (Long Papers)},  1401--1412.

\bibitem[\protect\citeauthoryear{Akbik, Blythe, and
  Vollgraf}{2018}]{akbik-etal-2018-contextual}
Akbik, A.; Blythe, D.; and Vollgraf, R.
\newblock 2018.
\newblock Contextual string embeddings for sequence labeling.
\newblock In {\em Proceedings of the 27th International Conference on
  Computational Linguistics},  1638--1649.
\newblock Santa Fe, New Mexico, USA: Association for Computational Linguistics.

\bibitem[\protect\citeauthoryear{Bodapati, Yun, and
  Al-Onaizan}{2019}]{bodapati-etal-2019-robustness}
Bodapati, S.; Yun, H.; and Al-Onaizan, Y.
\newblock 2019.
\newblock Robustness to capitalization errors in named entity recognition.
\newblock In {\em Proceedings of the 5th Workshop on Noisy User-generated Text
  (W-NUT 2019)},  237--242.
\newblock Hong Kong, China: Association for Computational Linguistics.

\bibitem[\protect\citeauthoryear{Brown and
  Coden}{2001}]{brown2001capitalization}
Brown, E.~W., and Coden, A.~R.
\newblock 2001.
\newblock Capitalization recovery for text.
\newblock In {\em Workshop on Information Retrieval Techniques for Speech
  Applications},  11--22.
\newblock Springer.

\bibitem[\protect\citeauthoryear{Chelba and Acero}{2006}]{chelba2006adaptation}
Chelba, C., and Acero, A.
\newblock 2006.
\newblock Adaptation of maximum entropy capitalizer: Little data can help a
  lot.
\newblock {\em Computer Speech \& Language} 20(4):382--399.

\bibitem[\protect\citeauthoryear{Chiu and Nichols}{2016}]{Chiu2016NamedER}
Chiu, J. P.~C., and Nichols, E.
\newblock 2016.
\newblock {Named Entity Recognition with Bidirectional LSTM-CNNs}.
\newblock {\em TACL} 4:357--370.

\bibitem[\protect\citeauthoryear{Clark \bgroup et al\mbox.\egroup
  }{2018}]{clark2018semi}
Clark, K.; Luong, M.-T.; Manning, C.~D.; and Le, Q.
\newblock 2018.
\newblock Semi-supervised sequence modeling with cross-view training.
\newblock In {\em Proceedings of the 2018 Conference on Empirical Methods in
  Natural Language Processing},  1914--1925.
\newblock Brussels, Belgium: Association for Computational Linguistics.

\bibitem[\protect\citeauthoryear{Collobert \bgroup et al\mbox.\egroup
  }{2011}]{collobert2011natural}
Collobert, R.; Weston, J.; Bottou, L.; Karlen, M.; Kavukcuoglu, K.; and Kuksa,
  P.
\newblock 2011.
\newblock Natural language processing (almost) from scratch.
\newblock {\em Journal of machine learning research} 12(Aug):2493--2537.

\bibitem[\protect\citeauthoryear{Coster and Kauchak}{2011}]{coster2011learning}
Coster, W., and Kauchak, D.
\newblock 2011.
\newblock Learning to simplify sentences using wikipedia.
\newblock In {\em Proceedings of the workshop on monolingual text-to-text
  generation},  1--9.
\newblock Association for Computational Linguistics.

\bibitem[\protect\citeauthoryear{Derczynski \bgroup et al\mbox.\egroup
  }{2017}]{derczynski2017results}
Derczynski, L.; Nichols, E.; van Erp, M.; and Limsopatham, N.
\newblock 2017.
\newblock Results of the {WNUT2017} shared task on novel and emerging entity
  recognition.
\newblock In {\em Proceedings of the 3rd Workshop on Noisy User-generated
  Text},  140--147.

\bibitem[\protect\citeauthoryear{Devlin \bgroup et al\mbox.\egroup
  }{2019}]{Devlin2018BERTPO}
Devlin, J.; Chang, M.-W.; Lee, K.; and Toutanova, K.
\newblock 2019.
\newblock {BERT}: Pre-training of deep bidirectional transformers for language
  understanding.
\newblock In {\em Proceedings of the 2019 Conference of the North {A}merican
  Chapter of the Association for Computational Linguistics: Human Language
  Technologies, Volume 1 (Long and Short Papers)},  4171--4186.
\newblock Minneapolis, Minnesota: Association for Computational Linguistics.

\bibitem[\protect\citeauthoryear{Finkel and Manning}{2009}]{FinkelMa09}
Finkel, J.~R., and Manning, C.~D.
\newblock 2009.
\newblock Nested named entity recognition.
\newblock In {\em Proc. of the Conference on Empirical Methods for Natural
  Language Processing (EMNLP)}.

\bibitem[\protect\citeauthoryear{Gardner \bgroup et al\mbox.\egroup
  }{2017}]{Gardner2017AllenNLP}
Gardner, M.; Grus, J.; Neumann, M.; Tafjord, O.; Dasigi, P.; Liu, N.~F.;
  Peters, M.; Schmitz, M.; and Zettlemoyer, L.~S.
\newblock 2017.
\newblock {AllenNLP: A Deep Semantic Natural Language Processing Platform}.

\bibitem[\protect\citeauthoryear{Hovy \bgroup et al\mbox.\egroup
  }{2006}]{HMPRW06}
Hovy, E.; Marcus, M.; Palmer, M.; Ramshaw, L.; and Weischedel, R.
\newblock 2006.
\newblock Ontonotes: The 90\% solution.
\newblock In {\em Proceedings of HLT/NAACL}.

\bibitem[\protect\citeauthoryear{Koehn \bgroup et al\mbox.\egroup
  }{2007}]{koehn2007moses}
Koehn, P.; Hoang, H.; Birch, A.; Callison-Burch, C.; Federico, M.; Bertoldi,
  N.; Cowan, B.; Shen, W.; Moran, C.; Zens, R.; et~al.
\newblock 2007.
\newblock Moses: Open source toolkit for statistical machine translation.
\newblock In {\em Proceedings of the 45th annual meeting of the association for
  computational linguistics companion volume proceedings of the demo and poster
  sessions},  177--180.

\bibitem[\protect\citeauthoryear{Kubala \bgroup et al\mbox.\egroup
  }{1998}]{kubala1998named}
Kubala, F.; Schwartz, R.; Stone, R.; and Weischedel, R.
\newblock 1998.
\newblock Named entity extraction from speech.
\newblock In {\em Proceedings of DARPA Broadcast News Transcription and
  Understanding Workshop},  287--292.
\newblock Citeseer.

\bibitem[\protect\citeauthoryear{Lample \bgroup et al\mbox.\egroup
  }{2016}]{Lample2016NeuralAF}
Lample, G.; Ballesteros, M.; Subramanian, S.; Kawakami, K.; and Dyer, C.
\newblock 2016.
\newblock Neural architectures for named entity recognition.
\newblock In {\em HLT-NAACL}.

\bibitem[\protect\citeauthoryear{Lita \bgroup et al\mbox.\egroup
  }{2003}]{lita2003truecasing}
Lita, L.~V.; Ittycheriah, A.; Roukos, S.; and Kambhatla, N.
\newblock 2003.
\newblock {tRuEcasIng}.
\newblock In {\em Proceedings of the 41st Annual Meeting on Association for
  Computational Linguistics-Volume 1},  152--159.
\newblock Association for Computational Linguistics.

\bibitem[\protect\citeauthoryear{Mayhew, Tsygankova, and
  Roth}{2019}]{mayhew-etal-2019-ner}
Mayhew, S.; Tsygankova, T.; and Roth, D.
\newblock 2019.
\newblock ner and pos when nothing is capitalized.
\newblock In {\em Proceedings of the 2019 Conference on Empirical Methods in
  Natural Language Processing and the 9th International Joint Conference on
  Natural Language Processing (EMNLP-IJCNLP)},  6257--6262.
\newblock Hong Kong, China: Association for Computational Linguistics.

\bibitem[\protect\citeauthoryear{Nebhi, Bontcheva, and
  Gorrell}{2015}]{nebhi2015restoring}
Nebhi, K.; Bontcheva, K.; and Gorrell, G.
\newblock 2015.
\newblock Restoring capitalization in tweets.
\newblock In {\em Proceedings of the 24th International Conference on World
  Wide Web},  1111--1115.
\newblock ACM.

\bibitem[\protect\citeauthoryear{Pennington, Socher, and
  Manning}{2014}]{Pennington2014GloveGV}
Pennington, J.; Socher, R.; and Manning, C.~D.
\newblock 2014.
\newblock Glove: Global vectors for word representation.
\newblock In {\em EMNLP}.

\bibitem[\protect\citeauthoryear{Peters \bgroup et al\mbox.\egroup
  }{2018}]{Peters2018DeepCW}
Peters, M.~E.; Neumann, M.; Iyyer, M.; Gardner, M.; Clark, C.; Lee, K.; and
  Zettlemoyer, L.~S.
\newblock 2018.
\newblock Deep contextualized word representations.
\newblock In {\em NAACL-HLT}.

\bibitem[\protect\citeauthoryear{Pradhan \bgroup et al\mbox.\egroup
  }{2012}]{pradhan2012conll}
Pradhan, S.; Moschitti, A.; Xue, N.; Uryupina, O.; and Zhang, Y.
\newblock 2012.
\newblock Conll-2012 shared task: Modeling multilingual unrestricted
  coreference in ontonotes.
\newblock In {\em Joint Conference on EMNLP and CoNLL-Shared Task},  1--40.
\newblock Association for Computational Linguistics.

\bibitem[\protect\citeauthoryear{Ramshaw and Marcus}{1996}]{RamshawMa96}
Ramshaw, L.~A., and Marcus, M.~P.
\newblock 1996.
\newblock Exploring the nature of transformation-based learning.
\newblock In Klavans, J., and Resnik, P., eds., {\em The Balancing Act:
  Combining Symbolic and Statistical Approaches to Language}.

\bibitem[\protect\citeauthoryear{Ratinov and Roth}{2009}]{RatinovRo09}
Ratinov, L., and Roth, D.
\newblock 2009.
\newblock Design challenges and misconceptions in named entity recognition.
\newblock In {\em Proc. of the Conference on Computational Natural Language
  Learning (CoNLL)}.

\bibitem[\protect\citeauthoryear{Ritter \bgroup et al\mbox.\egroup
  }{2011}]{ritter2011named}
Ritter, A.; Clark, S.; Etzioni, O.; et~al.
\newblock 2011.
\newblock Named entity recognition in tweets: an experimental study.
\newblock In {\em Proceedings of the conference on empirical methods in natural
  language processing},  1524--1534.
\newblock Association for Computational Linguistics.

\bibitem[\protect\citeauthoryear{Sang and
  Meulder}{2003}]{Sang2003IntroductionTT}
Sang, E. T.~K., and Meulder, F.~D.
\newblock 2003.
\newblock Introduction to the conll-2003 shared task: Language-independent
  named entity recognition.
\newblock In {\em CoNLL}.

\bibitem[\protect\citeauthoryear{Strauss \bgroup et al\mbox.\egroup
  }{2016}]{strauss2016results}
Strauss, B.; Toma, B.; Ritter, A.; De~Marneffe, M.-C.; and Xu, W.
\newblock 2016.
\newblock Results of the wnut16 named entity recognition shared task.
\newblock In {\em Proceedings of the 2nd Workshop on Noisy User-generated Text
  (WNUT)},  138--144.

\bibitem[\protect\citeauthoryear{Susanto, Chieu, and
  Lu}{2016}]{susanto2016learning}
Susanto, R.~H.; Chieu, H.~L.; and Lu, W.
\newblock 2016.
\newblock Learning to capitalize with character-level recurrent neural
  networks: an empirical study.
\newblock In {\em Proceedings of the 2016 Conference on Empirical Methods in
  Natural Language Processing},  2090--2095.

\bibitem[\protect\citeauthoryear{Wang, Knight, and
  Marcu}{2006}]{wang2006capitalizing}
Wang, W.; Knight, K.; and Marcu, D.
\newblock 2006.
\newblock Capitalizing machine translation.
\newblock In {\em Proceedings of the Human Language Technology Conference of
  the NAACL, Main Conference}.

\end{thebibliography}
\bibliographystyle{aaai}

\end{document}